\documentclass[11pt]{article}
\usepackage{csquotes}  %

\usepackage[preprint]{acl}

\usepackage{times}
\usepackage{latexsym}

\usepackage[T1]{fontenc}

\usepackage[utf8]{inputenc}

\usepackage{microtype}
\usepackage{amssymb}
\usepackage{inconsolata}

\usepackage{graphicx}
\usepackage{subcaption}  %

\title{CacheSpec: Finding the Sweet Spot for Small Models in Large Language Models}

\author{
Jingquan Chen$^{1,2,*}$ \quad
Jie Feng$^{2,*,\dagger}$ \quad
Jinghua Piao$^{3,2,\dagger}$ \quad
Shaogang Hu$^{1}$ \quad
Yong Li$^{3,2,\dagger}$ \\
$^{1}$University of Electronic Science and Technology of China \\
$^{2}$Zhongguancun Academy \\
$^{3}$Department of Electronic Engineering, BNRist, Tsinghua University
}

\begin{document}
\maketitle

\begin{abstract}
Large language models (LLMs) are increasingly used for program-aided reasoning, agentic decision making, and structured task execution, but these settings often incur substantial inference cost. Many such requests share similar computational structures while differing in variables, constraints, or contexts, creating opportunities for program-level caching. Since program caches need to reapply reusable computation logic to new requests, their key steps often involve lightweight and structured operations such as variable extraction, program binding, and generation acceleration, which are well suited for small models. We propose CacheSpec, an inference optimization framework centered on reusable program caches. The framework converts Program-of-Thoughts (PoT)-style programs from one-time reasoning artifacts into reusable cache objects, and reuses the same small model for two roles: semantic variable extraction on the cache-hit path and speculative drafting during target-LLM generation. Experiments on shopping-style request datasets, WebShop, Formula, and CodeTAT-QA show that CacheSpec reduces inference latency and improves effective cache reuse while preserving comparable or better task quality than existing caching and generation baselines, achieving up to about 3.1$\times$ latency speedup; in parallel serving experiments, it improves throughput by about 2.8$\times$ over PoT-style methods. These results suggest that the sweet spot for small models in large-model inference systems lies not in solving complex tasks independently, but in performing lightweight, structured, and verifiable auxiliary operations.
\end{abstract}

\setcounter{footnote}{0} %
\footnotetext[1]{Equal contribution, $^{\dagger}$Corresponding authors}

\section{Introduction}

Large Language Models (LLMs) have demonstrated strong capabilities in complex reasoning, code generation, tool use, and agentic task execution. This progress has been closely associated with scaling model size, training data, and computation \citep{brown2020language,kaplan2020scaling,hoffmann2022training}. Recent systems continue this trend through both large dense models and sparse Mixture-of-Experts architectures, such as Llama 3 and DeepSeek-V3 \citep{grattafiori2024llama,deepseek2024v3}. However, larger target LLMs also introduce higher inference latency, memory pressure, and serving cost. Many program-aided reasoning tasks, structured task executions, and agentic workflows involve related requests that differ in request-specific variables, constraints, or contexts, yet share similar computational structures, action templates, or variable relations. Therefore, instead of generating a new program for every request, the system can potentially reuse computation logic from previously generated programs.

We focus on where small models can provide value in this reusable-program setting. Unlike response-level caching, reusable program caching does not simply return a previous answer; rather, it must adapt existing program logic to new inputs through intermediate operations between the new request and the cached program. These operations are typically well-bounded, low-cost, and easier to constrain than full task solving. At the same time, small models are cheaper to invoke, faster to respond, and suitable for local structured tasks, making them a natural fit for such auxiliary operations rather than for independently solving the entire complex task.

Existing LLM inference optimization methods reduce inference cost from different directions. Program-aided reasoning methods, including Program-of-Thoughts (PoT) and Program-aided Language Models (PAL), translate natural-language problems into executable programs and improve reliability on computation-heavy tasks \citep{chen2023program,gao2023pal}. However, they typically require the target LLM to generate a new program for every request. More importantly, these one-time programs often entangle computation logic with request-specific variables, and may even hardcode concrete values or conditions into the generated code. Therefore, they are difficult to directly cache and reuse for new requests from the same task family. Cache-based reuse methods, such as GPTCache and GenCache, can reduce repeated inference, but their quality depends on whether cache matching and variable binding are reliable \citep{bang2023gptcache,chakraborty2025generative}. Speculative Decoding and Speculative Sampling accelerate a single target-model generation through draft-and-verify mechanisms, but they do not directly exploit computational homogeneity across requests \citep{leviathan2023fast,chen2023accelerating}.

Representative directions include PoT and PAL for program-aided reasoning, Speculative Decoding and Speculative Sampling for draft-and-verify acceleration, and GPTCache and GenCache for cache-based reuse. Table~\ref{tab:intro_comparison} summarizes these methods from the perspectives of task accuracy, speed, and resource utilization. PoT-style methods mainly improve accuracy on complex computation tasks, but often sacrifice speed and resource utilization. Speculative decoding methods mainly improve generation speed, while their resource-utilization benefit is conditional because they introduce an additional draft model. Cache-based reuse can improve speed and resource utilization, but its accuracy depends on the reliability of cache hits. In contrast, our goal is to organize program reasoning, cache reuse, and small-model acceleration within a unified inference framework.

\begin{table}[t]
\centering
\caption{
Qualitative comparison of representative inference optimization methods.
Resource Util. refers to reducing target-LLM computation.
}
\label{tab:intro_comparison}
\footnotesize
\setlength{\tabcolsep}{3pt}
\renewcommand{\arraystretch}{1.12}
\resizebox{\columnwidth}{!}{
\begin{tabular}{lccc}
\hline
\textbf{Method} & \textbf{Acc.} & \textbf{Speed} & \textbf{Resource Util.} \\
\hline
PoT & \checkmark & $\times$ & $\times$ \\
PAL & \checkmark & $\times$ & $\times$ \\
Speculative Decoding & -- & \checkmark & -- \\
Speculative Sampling & -- & \checkmark & -- \\
GPTCache & -- & \checkmark & \checkmark \\
GenCache & -- & \checkmark & \checkmark \\
CacheSpec & \checkmark & \checkmark & \checkmark \\
\hline
\end{tabular}
}
\end{table}

To address this gap, we propose CacheSpec (Program-Cache-SpecDec), an LLM inference optimization framework centered on reusable program caches. The key idea is to convert PoT-style programs from one-time reasoning artifacts into parameterized cache objects, thereby decoupling computation logic from request-specific data. Each cache entry contains a variable extraction template and a parameterized program: the template specifies which variables should be extracted from a new request, while the program stores reusable computation logic across requests. For cache-hit requests, the system uses a small model to extract semantic variables and binds them to the cached program for execution. For cache-miss requests and cache generation, the target LLM generates new templates and programs, while the same small model serves as a speculative drafter to reduce generation cost.

This design assigns the small model to lightweight, structured operations that are central to reusable program caching. Semantic variable extraction allows cached programs to adapt to new requests that are semantically related but structurally different. Speculative drafting reduces the cost of target-LLM calls that remain necessary during cache-miss inference and cache construction. In this way, the small model supports reusable program caching not by solving the full task independently, but by performing auxiliary operations that are easier to verify and cheaper to execute.

We evaluate CacheSpec on the shopping-style request datasets introduced by GenCache \citep{chakraborty2025generative}, WebShop, Formula from FinLoRA \citep{wang2025finlora}, and CodeTAT-QA from BizBench \citep{krumdick2024bizbench}, and further analyze its robustness under long contexts and parallel serving. On a financial reasoning task with stable computational structures, CacheSpec preserves PoT-style-level accuracy while achieving about 3.1$\times$ latency speedup. In parallel serving experiments, it achieves about 2.8$\times$ throughput improvement over PoT-style. These results suggest that the sweet spot for small models in large-model inference systems lies not in solving complex tasks independently, but in performing lightweight, structured, and verifiable auxiliary operations.\footnote{Our implementation and experimental artifacts are publicly available at \url{https://github.com/chenjqQAQ/CacheSpec}.} The main contributions are summarized as follows:

\begin{itemize}
    \item We propose CacheSpec, an LLM inference optimization framework centered on reusable program caches. The framework converts PoT-style programs into parameterized cache objects for cross-request reuse, decouples computation logic from request-specific data, and unifies program reasoning, cache reuse, and generation acceleration in one inference system.

    \item We design a dual-role small-model reuse mechanism. The same small model serves as a semantic variable extractor on the cache-hit path, enabling cached programs to adapt to structurally different but semantically related requests, and as a speculative drafter during target-LLM generation, reducing the cost of cache-miss inference and cache construction.

    \item We conduct extensive experiments across shopping-style request datasets, WebShop, Formula, and CodeTAT-QA, with additional analysis under long-context inputs and parallel serving. The results demonstrate the effectiveness of CacheSpec in reducing inference latency and improving cache reuse while maintaining competitive task quality.
\end{itemize}

\section{Related Work}

\paragraph{Program-aided reasoning.}
Program-aided reasoning improves LLM reliability on complex reasoning tasks by using explicit intermediate reasoning or executable programs \citep{wei2022chain,chen2023program,gao2023pal,yao2023react}. Chain-of-Thought prompting exposes intermediate reasoning steps, while Program of Thoughts and PAL translate natural-language problems into executable code that can be handled by external interpreters \citep{wei2022chain,chen2023program,gao2023pal}. ReAct further extends this idea to interactive settings by interleaving reasoning traces with task-specific actions \citep{yao2023react}. Although these methods improve accuracy, they usually generate a new program for each request, causing repeated latency and token cost when requests share the same computation pattern but differ only in variables or surface forms.

\paragraph{Speculative decoding.}
Speculative decoding reduces autoregressive generation latency through a draft-and-verify paradigm, where a smaller or faster model proposes candidate tokens and the target LLM verifies them in parallel \citep{leviathan2023fast,chen2023accelerating}. Recent variants improve this paradigm with different verification structures, drafting mechanisms, self-speculation, and lookahead decoding \citep{miao2024specinfer,cai2024medusa,li2024eagle,zhang2024draft,fu2024lookahead}. These methods accelerate single target-model generation, but they do not directly exploit reusable computation structures across related requests, and some variants require additional draft models, auxiliary heads, or model-specific modifications.

\paragraph{Cache-based reuse.}
Cache-based reuse reduces redundant LLM inference by reusing previous outputs, semantically similar queries, or model states \citep{bang2023gptcache,gill2024meancache,gim2024prompt,chakraborty2025generative}. Existing methods include response-level and semantic caches, as well as model-state caches that reuse prompt segments or attention states \citep{bang2023gptcache,gill2024meancache,gim2024prompt}. GenCache is closer to our setting because it studies generative reuse for structurally similar prompts and addresses the limitations of exact matching and purely semantic caching under request variations \citep{chakraborty2025generative}. However, program-aided reasoning requires not only matching similar requests, but also constructing reusable programs or templates and reliably extracting and binding variables from new inputs, making robust program-level reuse challenging.

\paragraph{Unified small-model reuse.}
PoT-style reasoning, speculative decoding, and GenCache-style caching optimize LLM inference from program generation, generation acceleration, and structural reuse, respectively \citep{chen2023program,leviathan2023fast,chen2023accelerating,chakraborty2025generative}. However, these mechanisms are often studied separately, leaving underexplored how they can be unified and where small models provide the highest leverage. We argue that PoT-style programs can serve not only as one-time reasoning artifacts but also as reusable cache objects for computation-oriented tasks. Under this view, small models can support both semantic variable extraction for reliable cache execution and speculative drafting for efficient cache construction.

\section{Method}

\subsection{Overview}

\begin{figure*}[t]
\centering
\includegraphics[width=\textwidth]{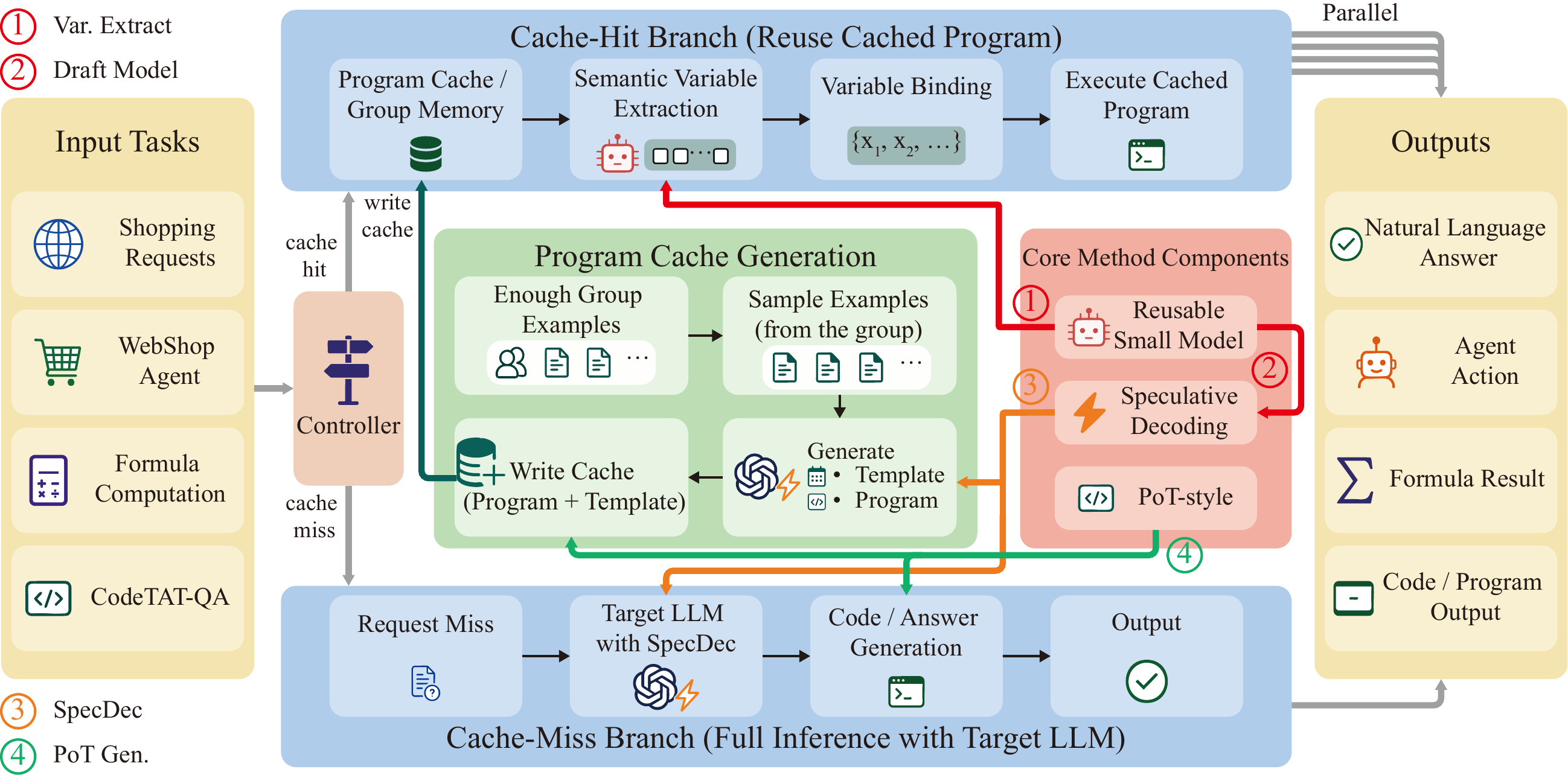}
\caption{
Overview of the proposed inference framework.
The numbered arrows denote:
(1) semantic variable extraction by the reusable small model,
(2) the same small model as the speculative drafter,
(3) SpecDec-accelerated target-LLM calls,
and (4) PoT-style program generation for answers and cache construction.
}
\label{fig:framework}
\end{figure*}

CacheSpec is an LLM inference optimization framework centered on reusable program caches, as shown in Figure~\ref{fig:framework}. For each input request, the framework first performs request routing. If a matched group has a valid program cache, the request enters the cache-hit branch, where a reusable small model extracts variables and an executable cached program produces the output. Otherwise, the request enters the cache-miss branch, where the target LLM generates an answer, code, or program, with Speculative Decoding (SpecDec) used to reduce generation cost. When a request group accumulates enough examples, the framework attempts to construct a new program cache for future reuse.

The framework follows a reuse-oriented design. Target-LLM-generated programs are converted from one-time reasoning artifacts into executable cache objects, and the same small model is reused for semantic variable extraction on the cache-hit path and speculative drafting during target-LLM generation. In this way, the small model supports reusable program caching through two lightweight operations: semantic variable extraction, which determines cache usability, and speculative drafting, which reduces the cost of remaining target-LLM generation.

\subsection{Routing and Program Cache}

Because many computation-oriented requests share executable structures, the framework routes each request before invoking the target LLM. Let $\mathcal{G}=\{G_1,G_2,\ldots,G_n\}$ denote the existing request groups. For an input request $x$, we use a semantic encoder $E(\cdot)$ to obtain its embedding $\mathbf{e}_x=E(x)$. We use \texttt{all-MiniLM-L6-v2} as the semantic encoder for semantic similarity search \citep{reimers2019sentencebert,wang2020minilm}. The system computes a semantic matching score $S(x,G_i)$ between $x$ and each group $G_i$, where $S(\cdot)$ uses embedding similarity.

The system selects the group with the highest matching score, $i^\ast=\arg\max_i S(x,G_i)$. If $S(x,G_{i^\ast})<\tau$, where $\tau$ denotes the semantic matching threshold, the request is sent to the cache-miss branch. Otherwise, the system checks whether $G_{i^\ast}$ has a valid program cache.

We define a program cache entry as $C_i=(T_i,P_i)$. Here, $T_i$ denotes a variable extraction template that specifies which variables the small model should extract and what format they should follow. $P_i$ denotes an executable program that takes the extracted variables as input and produces the final output. If group $G_{i^\ast}$ does not have a cache entry, the request enters the cache-miss branch; otherwise, it enters the cache-hit branch.

\subsection{Cache-Hit Branch}

A cached program can be reused only when the system correctly identifies and binds the required variables in a new request. Since surface-pattern-based extraction in GenCache-style reuse may fail for semantically similar but structurally different requests \citep{chakraborty2025generative}, we use the small model for semantic variable extraction to improve the robustness of cached-program reuse.

For a matched cache entry $C_{i^\ast}=(T_{i^\ast},P_{i^\ast})$, the system feeds the input request $x$ and the extraction template $T_{i^\ast}$ into the reusable small model $M_s$, which outputs a structured variable set $\mathbf{v}=M_s(x,T_{i^\ast})$. The variable set $\mathbf{v}$ may contain domain-specific entities, constraints, or numerical parameters depending on the template. The system then checks whether $\mathbf{v}$ contains all required fields and satisfies the expected types and formats. If the check fails, the request falls back to the cache-miss branch.

If the extraction is valid, the system binds $\mathbf{v}$ to the executable program $P_{i^\ast}$ and returns $y=P_{i^\ast}(\mathbf{v})$. Thus, the small model performs a lightweight but high-leverage auxiliary task: it extracts the variables required for safe program execution, while the cached program preserves the reusable computation logic. The cache-hit path can also batch variable extraction under concurrent serving; implementation details are provided in Appendix~\ref{app:cache_hit_details}.

\subsection{Cache-Miss Branch}

The cache-miss branch handles requests that cannot be safely served by the Program Cache, including cases where no matching group exists, the matched group has no generated cache, variable extraction fails, or the task requires new reasoning. For an input request $x$ in this branch, the target LLM $M_t$ generates an answer, code, or executable program. We use the same small model $M_s$ as the speculative drafter during target-LLM invocation. Following the draft-and-verify mechanism in SpecDec, the small model proposes candidate tokens or steps, and the target LLM verifies, corrects, or continues the generation \citep{leviathan2023fast,chen2023accelerating}.

The output form depends on the task. For computation-oriented or structured reasoning tasks, the target LLM can generate code or a program whose execution produces the final result. For tasks that do not require execution, it can directly generate a natural-language answer or an agent action. This fallback branch preserves the reasoning capability of the target LLM while reducing the cost of necessary target-LLM calls through speculative drafting.

\subsection{Cache Generation}

The framework generates program caches so that target-LLM outputs do not remain one-time reasoning artifacts. For a request group $G_i$, let $\mathcal{D}_i=\{(x_j,y_j)\}_{j=1}^{m_i}$ denote its accumulated examples, where $x_j$ is a request and $y_j$ is the corresponding output. Each group has a dynamic cache generation threshold $\nu_i$. When $m_i\geq\nu_i$, the system triggers cache generation for group $G_i$.

During one cache generation attempt, the system samples examples from $\mathcal{D}_i$ for generation and validation. The generation examples prompt the target LLM to infer the shared structure of the group and generate a cache entry $C_i=(T_i,P_i)$. The target-LLM call that generates $T_i$ and $P_i$ also uses $M_s$ as the speculative drafter. The generated cache entry is then evaluated on validation examples: for each validation request, the small model extracts variables according to $T_i$, and the program $P_i$ executes with the extracted variables. Let $r_i$ denote the validation pass rate and $\alpha$ denote the required pass-rate threshold.

If $r_i\geq\alpha$, the system writes $C_i=(T_i,P_i)$ into the Program Cache, so future requests routed to $G_i$ can reuse this cache entry through the cache-hit branch. If cache generation fails, the system retries up to $R$ times and uses failure summaries as reflection for subsequent attempts. If repeated failures indicate that the group lacks a stable shared computational structure, the group is marked as uncacheable and future requests in this group continue to use the cache-miss branch.

For groups that remain potentially cacheable, we adopt exponential backoff to avoid repeatedly spending target-LLM calls on groups that are not ready for reliable cache construction. If group $G_i$ fails to generate a valid cache under the current threshold $\nu_i$, the system increases the next trigger threshold by updating $\nu_i \leftarrow \beta\nu_i$, where $\beta>1$ denotes the backoff coefficient. Detailed procedures for cache validation, retry, reflection, uncacheable-group marking, and backoff are provided in Appendix~\ref{app:cache_generation_details}.

\section{Experiments}

\subsection{Experimental Setup}

We implement an API-based evaluation framework that wraps different inference methods as OpenAI-compatible \texttt{/v1/chat/completions} services. Each method receives the same input format, and a unified runner sends requests, records request-level logs, saves cache states and task outputs, and computes evaluation metrics with the same scripts. This setup allows us to compare Direct LLM, PoT-style code generation, ExactCache, GPTCache, GenCache, CacheSpec, and SpecDec variants under a shared serving and evaluation pipeline.

We use Qwen3-32B as the target LLM and Qwen3-1.7B as the reusable small model \citep{yang2025qwen3}. We use Qwen3-1.7B without task-specific fine-tuning. The small model serves two roles: semantic variable extraction in the cache-hit branch and speculative drafting during target LLM invocation. For request-group matching, we use \texttt{all-MiniLM-L6-v2} as the semantic encoder, following sentence embedding and compact transformer models for semantic similarity search \citep{reimers2019sentencebert,wang2020minilm}.

For methods that support SpecDec, we evaluate both non-SpecDec and SpecDec variants. The non-SpecDec variants directly invoke Qwen3-32B for generation, while the SpecDec variants use Qwen3-1.7B as the draft model. The main tables report representative +SpecDec variants for readability, while the complete +SpecDec results are provided in Appendix~\ref{app:full_specdec_results}. Except for the parallel experiments, all experiments run in a single-request setting. For hyperparameters related to semantic matching, cache generation, validation, retry, and backoff, we keep the settings fixed within each task to ensure fair comparison across methods.

Additional revision experiments on model choice, cold-start scale, routing-threshold sensitivity, semantic variable extraction, cache-error propagation, repeated-run stability, and key hyperparameters are reported in Appendix~\ref{app:revision_experiments}.

\subsection{Datasets}

We evaluate on five groups of tasks. We use the Shopping-Full and Shopping-Struct request datasets introduced by GenCache \citep{chakraborty2025generative}. Shopping-Full contains 10,136 shopping-style requests, while Shopping-Struct contains 10,000 structurally perturbed requests that preserve the same semantic intent but change the prompt surface form. WebShop evaluates agent-style shopping behavior in a simulated e-commerce environment with real product data and natural-language goals \citep{yao2022webshop}. For financial reasoning, we use Formula from FinLoRA \citep{wang2025finlora} and CodeTAT-QA from BizBench \citep{krumdick2024bizbench}. Formula focuses on formula construction and calculation over XBRL financial data, whereas CodeTAT-QA involves more diverse table-and-text reasoning. We further construct length-controlled Formula variants with average input lengths near 1K, 2K, 4K, and 8K tokens to test robustness under long contexts. Detailed dataset construction and filtering procedures are provided in Appendix~\ref{app:datasets}.

\subsection{Baselines and Metrics}

We compare Direct LLM, PoT-style code generation, ExactCache, GPTCache, GenCache \citep{chakraborty2025generative}, and CacheSpec. For GenCache, we use a faithful reproduction of the original regex/matcher-based program-cache pipeline and adapt only task-specific input/output interfaces. For methods that involve target-LLM generation, we also evaluate +SpecDec variants. PoT-style is used only for Formula and CodeTAT-QA, where executable program generation is a natural baseline. For Shopping-Full, Shopping-Struct, Formula, and CodeTAT-QA, we report answer accuracy; for WebShop, we report average reward. Across tasks, Lat. denotes average end-to-end request latency. For cache-based methods, Hit denotes cache hit rate and Hit Acc. denotes the accuracy of cache-hit outputs. For parallel experiments, we additionally report throughput and wall-clock time per request. Additional baseline definitions and metric details are provided in Appendix~\ref{app:baselines_metrics}.

\subsection{Main Results}

Table~\ref{tab:main_shopping_webshop} and Table~\ref{tab:main_bizbench} report the main results. Overall, CacheSpec improves the latency--reuse trade-off while maintaining competitive task quality. On Shopping-Full and Shopping-Struct, it achieves high Hit and Hit Acc. with lower latency than Direct LLM, showing that small-model variable extraction enables reliable program reuse under request variations. GPTCache also obtains high Hit but poor quality, indicating that full-prompt semantic similarity alone is insufficient for safe reuse, while GenCache has limited reuse under structural perturbations. On financial reasoning tasks, CacheSpec improves accuracy over Direct LLM through PoT-style executable reasoning and reduces latency through program reuse. More detailed analysis is provided in Appendix~\ref{app:detailed_main_results}.

\begin{table*}[t]
\centering
\caption{
Main results on Shopping-Full, Shopping-Struct, and WebShop.
Acc. denotes accuracy, Reward denotes average WebShop reward, Lat. denotes average end-to-end latency, and Hit/Hit Acc. denote cache hit rate and cache-hit accuracy.
}
\label{tab:main_shopping_webshop}
\footnotesize
\setlength{\tabcolsep}{3pt}
\renewcommand{\arraystretch}{1.08}
\resizebox{\textwidth}{!}{
\begin{tabular}{lccccccccccc}
\hline
\textbf{Method}
& \multicolumn{4}{c}{\textbf{Shopping-Full}}
& \multicolumn{4}{c}{\textbf{Shopping-Struct}}
& \multicolumn{3}{c}{\textbf{WebShop}} \\
\cline{2-5}\cline{6-9}\cline{10-12}
& Acc. & Lat. & Hit & Hit Acc.
& Acc. & Lat. & Hit & Hit Acc.
& Reward & Lat. & Hit \\
\hline
Direct LLM
& 92.71 & 0.915 & -- & --
& 97.93 & 0.913 & -- & --
& 0.5728 & 1.4070 & -- \\
Direct LLM + SpecDec
& \textbf{92.73} & 0.419 & -- & --
& \textbf{97.94} & 0.475 & -- & --
& 0.5762 & 1.1364 & -- \\
ExactCache
& 92.72 & 0.972 & 0.00 & --
& 97.93 & 0.971 & 0.00 & --
& \textbf{0.5832} & 1.4515 & 0.00 \\
GPTCache
& 8.11 & 0.108 & 91.80 & 0.02
& 17.18 & 0.150 & 89.17 & 7.13
& 0.0531 & 0.1942 & 89.33 \\
GenCache
& 78.51 & 0.197 & 94.89 & 77.68
& 96.91 & 3.252 & 0.10 & 0.00
& 0.5600 & 0.7264 & 70.52 \\
CacheSpec
& 91.91 & 0.256 & \textbf{96.80} & 91.89
& 96.65 & 0.249 & \textbf{98.03} & 96.62
& 0.5763 & 0.6598 & 74.22 \\
CacheSpec + SpecDec
& 92.15 & \textbf{0.214} & 96.35 & \textbf{92.13}
& 96.66 & \textbf{0.228} & \textbf{98.03} & \textbf{96.64}
& 0.5760 & \textbf{0.5797} & 74.22 \\
\hline
\end{tabular}
}
\end{table*}

The financial reasoning tasks further show the advantage of converting PoT-style programs into reusable cache objects. On Formula, PoT-style improves Acc. from 75.92 to 94.69 but increases Lat. from 0.237s to 2.015s. CacheSpec + SpecDec preserves PoT-style-level accuracy, achieving 94.19 Acc., while reducing Lat. to 0.648s with 87.16 Hit and 93.76 Hit Acc. Compared with the reproduced GenCache baseline, which reaches 94.52 Acc. but has 0.00 Hit on Formula, CacheSpec obtains effective cache reuse rather than relying mainly on fallback generation. On CodeTAT-QA, where group-level computation patterns are less stable, CacheSpec still serves more than half of the requests through cache, but with a slight accuracy drop, indicating that reusable program caching is most effective when tasks contain stable computation structures.

\begin{table*}[t]
\centering
\caption{
Main results on Formula and CodeTAT-QA.
Acc. denotes answer accuracy, Lat. denotes average end-to-end latency, and Hit/Hit Acc. denote cache hit rate and cache-hit accuracy.
}
\label{tab:main_bizbench}
\footnotesize
\setlength{\tabcolsep}{4pt}
\renewcommand{\arraystretch}{1.08}
\begin{tabular}{lcccccccc}
\hline
\textbf{Method}
& \multicolumn{4}{c}{\textbf{Formula}}
& \multicolumn{4}{c}{\textbf{CodeTAT-QA}} \\
\cline{2-5}\cline{6-9}
& Acc. & Lat. & Hit & Hit Acc.
& Acc. & Lat. & Hit & Hit Acc. \\
\hline
Direct LLM
& 75.92 & 0.237 & -- & --
& 49.17 & 0.258 & -- & -- \\
Direct LLM + SpecDec
& 75.80 & 0.314 & -- & --
& 49.24 & 0.472 & -- & -- \\
PoT-style
& \textbf{94.69} & 2.015 & -- & --
& 71.53 & 2.447 & -- & -- \\
PoT-style + SpecDec
& 94.56 & 1.564 & -- & --
& 71.69 & 1.880 & -- & -- \\
ExactCache
& 94.64 & 2.056 & 0.00 & --
& \textbf{71.98} & 2.550 & 0.08 & 100.00 \\
GPTCache
& 94.56 & 0.858 & 63.45 & 93.69
& 42.54 & 0.485 & 83.03 & 35.73 \\
GenCache
& 94.52 & 3.678 & 0.00 & --
& 71.69 & 3.232 & 0.00 & -- \\
CacheSpec
& 94.19 & 0.847 & 84.14 & 93.54
& 71.25 & 1.849 & 56.05 & 68.00 \\
CacheSpec + SpecDec
& 94.19 & \textbf{0.648} & \textbf{87.16} & \textbf{93.76}
& 70.41 & \textbf{1.648} & \textbf{57.48} & \textbf{71.79} \\
\hline
\end{tabular}
\end{table*}

SpecDec mainly helps when target-LLM generation is long enough to amortize draft-and-verify overhead, such as PoT-style code generation, cache-miss inference, and cache construction. This supports our design choice of reusing the same small model not as a direct replacement for the target LLM, but as both a semantic variable extractor and a speculative drafter.

\subsection{Ablation Study}

Table~\ref{tab:ablation_formula} reports the ablation study on Formula, the clearest setting for analyzing program-aided reasoning, cache generation, and cached-program execution.

\begin{table}[t]
\centering
\caption{
Ablation study on Formula.
Cache Gen. reports successful cache generations over attempted cache generations.
}
\label{tab:ablation_formula}
\footnotesize
\setlength{\tabcolsep}{2pt}
\renewcommand{\arraystretch}{1.08}
\resizebox{\columnwidth}{!}{
\begin{tabular}{lccccc}
\hline
\textbf{Method} & Acc. & Lat. & Hit & Hit Acc. & Cache Gen. \\
\hline
Direct LLM
& 75.92 & 0.237 & -- & -- & -- \\
PoT-style
& \textbf{94.69} & 2.015 & -- & -- & -- \\
PoT-style + SpecDec
& 94.56 & 1.564 & -- & -- & -- \\
GenCache
& 94.52 & 3.678 & 0.00 & -- & 5/133 \\
CacheSpec
& 94.19 & 0.847 & 84.14 & 93.54 & 23/38 \\
CacheSpec + SpecDec
& 94.19 & \textbf{0.648} & \textbf{87.16} & \textbf{93.76} & \textbf{24/38} \\
\hline
\end{tabular}
}
\end{table}

The ablation results show how the three components support CacheSpec. PoT-style code generation provides accurate executable reasoning but incurs high latency. GenCache reaches PoT-style-level accuracy but generates few valid caches and obtains 0.00 Hit, meaning that its accuracy mainly comes from fallback code generation. In contrast, CacheSpec increases Cache Gen. from 5/133 to 23/38, obtains 84.14 Hit with 93.54 Hit Acc., and reduces Lat. to 0.847s. Adding SpecDec further reduces Lat. to 0.648s. These results show that small-model semantic extraction turns generated programs into reusable caches, while speculative drafting reduces the remaining target-LLM cost.

\subsection{Parallel Performance}

\begin{figure*}[!t]
\centering
\includegraphics[width=0.82\textwidth]{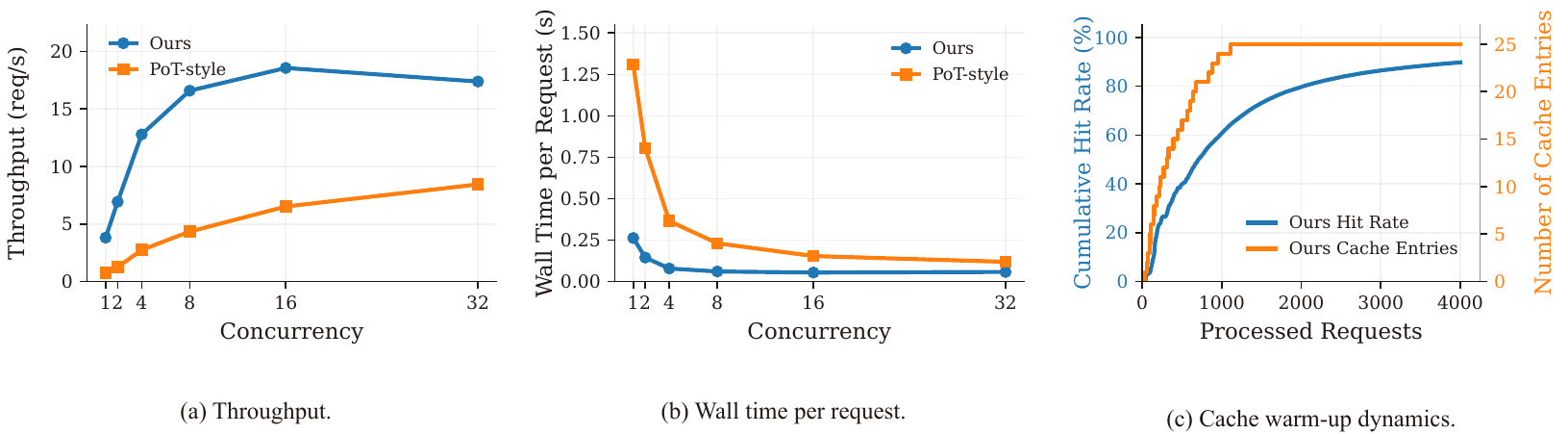}
\vspace{-0.6em}
\caption{
Parallel performance and cache warm-up on Formula.
(a) Throughput under different concurrency levels.
(b) Wall-clock time per request under different concurrency levels.
(c) Cache warm-up dynamics at concurrency 8.
}
\label{fig:parallel_indicators}
\vspace{-1.0em}
\end{figure*}

We evaluate parallel performance on Formula by comparing CacheSpec with PoT-style. As shown in Figure~\ref{fig:parallel_indicators}, CacheSpec achieves higher throughput and lower wall-clock time per request across concurrency levels. At concurrency 16, it reaches 18.56 req/s, compared with 6.51 req/s for PoT-style, giving about 2.85$\times$ throughput improvement. This advantage comes from replacing repeated target-LLM program generation with variable extraction and cached-program execution after cache warm-up. Detailed throughput curves and cache warm-up dynamics are provided in Appendix~\ref{app:parallel_performance}.

\subsection{Length Robustness}

\begin{figure}[t]
\centering
\begin{subfigure}[t]{0.48\columnwidth}
    \centering
    \includegraphics[width=\linewidth]{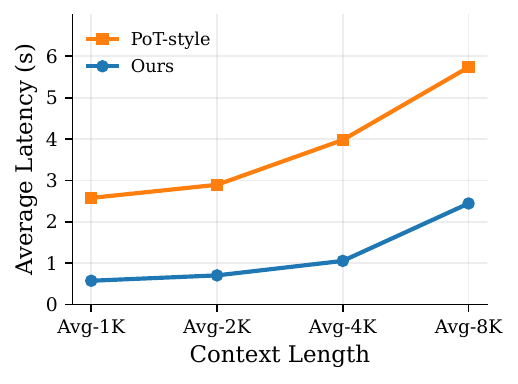}
    \caption{Avg. latency.}
    \label{fig:length_latency}
\end{subfigure}
\hfill
\begin{subfigure}[t]{0.48\columnwidth}
    \centering
    \includegraphics[width=\linewidth]{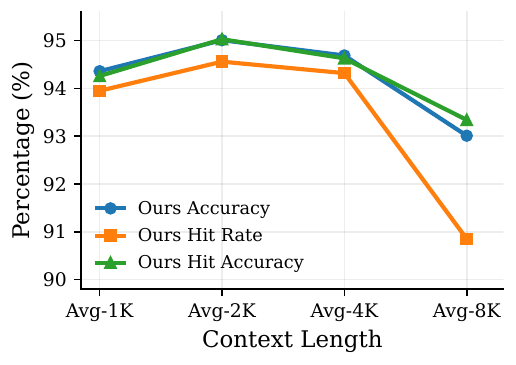}
    \caption{Quality.}
    \label{fig:length_quality}
\end{subfigure}
\vspace{-0.6em}
\caption{
Length robustness on Formula.
}
\label{fig:length_formula}
\vspace{-1.0em}
\end{figure}

Figure~\ref{fig:length_formula} evaluates whether program caching remains effective as the input context becomes longer. CacheSpec consistently outperforms PoT-style across Avg-1K to Avg-8K Formula settings, with latency speedups of 4.49$\times$, 4.11$\times$, 3.77$\times$, and 2.35$\times$, respectively. Although the speedup decreases at Avg-8K, CacheSpec still maintains 93.01\% accuracy, 90.85\% cache hit rate, and 93.34\% cache-hit accuracy, showing that reusable program caching remains effective under long contexts when the task has stable formula-level computation structures. Detailed analysis is provided in Appendix~\ref{app:length_robustness}.

\section{Conclusion}

We present an LLM inference optimization framework centered on reusable program caches. Instead of treating program-aided reasoning, cache reuse, and speculative decoding as separate mechanisms, our framework organizes them around a unified reuse principle: programs generated by the target LLM become executable cache objects, and the small model is reused for both semantic variable extraction and speculative drafting. This design allows the system to avoid repeated target-LLM program generation on cache-hit requests while reducing the cost of necessary target-LLM calls during cache-miss inference and cache construction.

Experiments on shopping-style request datasets, WebShop, Formula, and CodeTAT-QA show that our framework reduces inference latency and improves effective cache reuse while preserving competitive task quality compared with exact matching, semantic response caching, and generative program caching baselines. The results further show that reusable program caches remain effective under longer contexts and concurrent requests when tasks contain stable computation patterns. Overall, our findings suggest that the sweet spot for small models in large-model inference systems lies not in solving complex tasks independently, but in performing lightweight, structured, and verifiable auxiliary operations.

\section*{Limitations}

CacheSpec has three main limitations. First, it is most effective when requests share reusable structured patterns. Although small-model variable extraction improves reuse for semantically similar but structurally different requests, the framework remains less suitable for free-form dialogue, creative writing, or tasks with high prompt diversity. This is consistent with prior generative caching work, which also scopes caching to structurally similar repetitive tasks rather than free-form chatbot interactions \citep{chakraborty2025generative}.

Second, the method depends on both the target LLM and the small model. The target LLM must induce a valid extraction template and executable program from grouped examples, while the small model must reliably extract variables from new requests. If the group pattern is unstable, the generated program is incomplete, or the input contains long contexts and ambiguous variables, cache generation or cache-hit execution can become unreliable.

Third, program caching may propagate errors when the system constructs an incorrect cache or routes a request to an unsuitable cache entry. Prior work also notes this risk and recommends strict guardrails for generative caching \citep{chakraborty2025generative}. We reduce such failures with validation, fallback execution, failure reflection, exponential backoff, and uncacheable-group marking, but these mechanisms cannot eliminate all erroneous reuse.

\bibliography{custom}

\appendix
\section{Additional Experimental Details}
\label{sec:appendix}

\subsection{Cache-Hit Execution Details}
\label{app:cache_hit_details}

The cache-hit branch is designed to reuse cached programs without invoking the target LLM. Given a matched cache entry $C_{i^\ast}=(T_{i^\ast},P_{i^\ast})$, the small model receives the input request $x$ together with the variable extraction template $T_{i^\ast}$. The template specifies the required variable names, expected types, and output format. The small model then returns a structured variable set $\mathbf{v}=M_s(x,T_{i^\ast})$.

Before executing the cached program, the system validates the extracted variable set. The validation checks whether all required fields specified by $T_{i^\ast}$ are present, whether the extracted values satisfy the expected types and formats, and whether they can be safely bound to the program inputs. If any required field is missing, if a type constraint is violated, or if the output format cannot be parsed, the extraction is marked as invalid and the request falls back to the cache-miss branch.

When validation succeeds, the system binds $\mathbf{v}$ to the executable program $P_{i^\ast}$ and returns $y=P_{i^\ast}(\mathbf{v})$. This design separates semantic variable extraction from deterministic program execution: the small model handles the flexible natural-language interface, while the cached program preserves the reusable computation logic.

The cache-hit path also supports concurrent serving. Multiple variable extraction requests can be batched and sent to the small model together. After extraction, cached programs are executed independently for different requests. This batching reduces small-model overhead under concurrent workloads, while program execution avoids repeated target-LLM calls. This implementation is the basis for the parallel serving behavior analyzed in the main experiments.

\subsection{Cache Generation and Failure Handling Details}
\label{app:cache_generation_details}

For each request group $G_i$, the system maintains an accumulated example set $\mathcal{D}_i=\{(x_j,y_j)\}_{j=1}^{m_i}$ and a dynamic cache generation threshold $\nu_i$. When $m_i\geq\nu_i$, the group becomes eligible for cache generation. In each cache generation attempt, the system samples examples from $\mathcal{D}_i$ and splits them into generation examples and validation examples. The generation examples are used to prompt the target LLM to infer the shared computation pattern of the group and produce a cache entry $C_i=(T_i,P_i)$.

The generated template $T_i$ should describe the variables that must be extracted from future requests, including their names, expected types, and output structure. The generated program $P_i$ should implement the reusable computation logic shared by the group and take the extracted variables as input. The target-LLM call used for cache generation is also accelerated by SpecDec, where the same small model $M_s$ acts as the speculative drafter.

After a candidate cache entry is generated, the system evaluates it on validation examples. For each validation request, the small model extracts variables according to $T_i$, the program $P_i$ executes with the extracted variables, and the system checks whether the output satisfies the expected answer or output format. The validation pass rate is denoted as $r_i$. If $r_i\geq\alpha$, where $\alpha$ is the required pass-rate threshold, the cache entry is accepted and written into the Program Cache.

If validation fails, the system retries cache generation up to $R$ times. After each failed attempt, the target LLM summarizes the failure reasons. Typical failure reasons include incomplete variable definitions, inconsistent program logic, incorrect output format, invalid variable binding, or poor coverage of the validation examples. The failure summary is used as reflection to guide the next cache generation attempt, so later attempts can revise the extraction template or program logic according to previous errors.

If all $R$ attempts fail, the system records the failure summaries and reuses them when the group triggers cache generation again. When repeated failures indicate that the group lacks a stable shared computational structure, the group is marked as uncacheable. For an uncacheable group, the system disables future cache generation attempts and sends future requests in the group to the cache-miss branch. This prevents the framework from repeatedly spending target-LLM calls on groups that are unlikely to yield reliable program caches.

For groups that are not marked as uncacheable but still fail to produce a valid cache, the system applies exponential backoff. If group $G_i$ fails to generate a valid cache under the current threshold $\nu_i$, the next trigger threshold is increased as
\[
\nu_i \leftarrow \beta\nu_i,
\]
where $\beta>1$ is the backoff coefficient. This mechanism allows difficult groups to accumulate more examples before the next cache generation attempt and reduces wasted target-LLM computation during repeated unsuccessful cache construction.

\subsection{Dataset Construction Details}
\label{app:datasets}

\paragraph{Shopping request datasets.}
We use the Shopping-Full and Shopping-Struct request datasets introduced by GenCache \citep{chakraborty2025generative}. Shopping-Full contains 10,136 shopping-style requests. Each example includes a user instruction, the complete prompt sent to the model, and a reference action or answer. This dataset is used to evaluate whether different caching methods can reuse recurring shopping-style request patterns under the original prompt format.

Shopping-Struct is designed by GenCache to test robustness under structural perturbations. Starting from Shopping-Full, GenCache constructs rewritten versions that modify surface realization, field order, prompt format, or partial descriptions while aiming to preserve the same user intent. The dataset retains rewritten examples whose answers are judged to be equivalent to the original reference answer. This setting evaluates whether a cache method can reuse computation across prompts that are semantically equivalent but structurally different.

Table~\ref{tab:shopping_struct_example} gives a real example from Shopping-Struct. The rewritten instruction changes the surface form of the request, but the search intent and price constraint remain unchanged. The example is retained because the generated action is equivalent to the original reference action.

\begin{table}[t]
\centering
\caption{
Example of a Shopping-Struct perturbation.
}
\label{tab:shopping_struct_example}
\scriptsize
\setlength{\tabcolsep}{3pt}
\renewcommand{\arraystretch}{1.05}
\resizebox{\columnwidth}{!}{
\begin{tabular}{p{0.25\columnwidth}p{0.68\columnwidth}}
\hline
\textbf{Field} & \textbf{Example} \\
\hline
Source instruction
& i want a x-large and machine washable lazy tuxedo t-shirt, and price lower than 261 dollars \\
Rewritten instruction
& Could you help me find a lazy tuxedo t-shirt in x-large, machine washable, and under 261 dollars? \\
Reference action
& \texttt{Search(keywords=lazy tuxedo t-shirt x-large machine washable, max\_price=261)} \\
Equivalence check
& Qwen3-32B judges that the rewritten request preserves the same shopping intent and the same maximum-price constraint. \\
\hline
\end{tabular}
}
\end{table}

\paragraph{WebShop.}
WebShop is a simulated e-commerce environment with real product data and crowd-sourced instructions \citep{yao2022webshop}. Each episode requires an agent to search, inspect products, navigate pages, and select an item according to a natural-language goal. We evaluate 50 episodes using fixed product data, search indices, and task goals. Since WebShop outputs are mainly search, click, navigation, and selection actions, we do not evaluate PoT-style code generation on this task. We compare Direct LLM, ExactCache, GPTCache, GenCache, CacheSpec, and applicable SpecDec variants.

\paragraph{Financial reasoning tasks.}
We use Formula from FinLoRA \citep{wang2025finlora} and CodeTAT-QA from BizBench \citep{krumdick2024bizbench}. Formula contains Formula Construction and Formula Calculation tasks. These tasks are based on financial formulas and XBRL financial data: Formula Construction requires constructing financial formulas from relevant information, while Formula Calculation requires substituting numerical values into formulas and computing the final result. Formula is well suited for evaluating reusable program caches, because requests from the same formula group often share computation logic but differ in variable values. The original Formula dataset contains relatively few examples, so we expand it by using each original example as a seed and modifying non-critical fields, rephrasing questions, and changing input order while preserving the core formula logic and final answer. We use the locally deployed Qwen3-32B model without SpecDec to filter examples whose answers remain consistent with the seed.

Table~\ref{tab:formula_example} shows a real Formula example after rewriting. The example belongs to the Modigliani--Miller Theorem group. Although the question is rephrased, the underlying computation is unchanged: the model must substitute the unlevered firm value, tax rate, and debt into $V_L = V_U + T_cD$ and return the same numerical answer. Such examples allow us to test whether a cache method can learn a reusable computation pattern for a formula group rather than memorize a single prompt.

\begin{table}[t]
\centering
\caption{
Example of a Formula variant.
}
\label{tab:formula_example}
\scriptsize
\setlength{\tabcolsep}{3pt}
\renewcommand{\arraystretch}{1.05}
\resizebox{\columnwidth}{!}{
\begin{tabular}{p{0.25\columnwidth}p{0.68\columnwidth}}
\hline
\textbf{Field} & \textbf{Example} \\
\hline
Formula group
& Modigliani--Miller Theorem \\
Formula
& $V_L = V_U + T_cD$ \\
Seed question
& Assuming a company without any debt is valued at \$950{,}000, what would the value be should it decide to leverage \$150{,}000 at a 32\% corporate tax rate? \\
Rewritten question
& Compute the value of a levered firm given an unlevered value of \$950{,}000, debt of \$150{,}000, and a corporate tax rate of 32\%. \\
Reference answer
& 998000.0 \\
\hline
\end{tabular}
}
\end{table}

CodeTAT-QA is a more diverse financial table-and-text question answering task. Because the original data has weaker group-level regularities, we first cluster the dataset with \texttt{all-MiniLM-L6-v2} sentence embeddings and semantic similarity \citep{reimers2019sentencebert,wang2020minilm}. We then select large clusters to form a cacheable subset and randomly shuffle the selected examples. This construction tests whether reusable program caches remain useful when the task contains more complex financial reasoning and less regular request groups.

\paragraph{Length-controlled Formula datasets.}
To evaluate robustness under longer contexts, we construct four Formula variants with average input lengths near 1K, 2K, 4K, and 8K tokens. We add extra context, table fields, irrelevant text, or weakly related financial descriptions while preserving the original question logic, formula structure, and final answer. These datasets allow us to compare how Direct LLM, PoT-style code generation, and CacheSpec behave as context length increases. PoT-style methods must generate a new program over the full input context for every request, whereas cache-hit requests in CacheSpec only require variable extraction and cached program execution.

\subsection{Baseline and Metric Details}
\label{app:baselines_metrics}

\paragraph{Baselines.}
Direct LLM directly invokes the target model to generate the final answer. PoT-style denotes a program-generation-and-execution baseline, following the general idea of program-aided reasoning methods such as Program-of-Thoughts and PAL \citep{chen2023program,gao2023pal}. In this baseline, the target LLM generates executable code or a program, and the system executes it to obtain the final answer. We apply PoT-style only to Formula and CodeTAT-QA, where program execution is a natural fit.

ExactCache reuses cached responses only when the complete prompt exactly matches a previous prompt. GPTCache performs semantic response caching based on prompt embeddings, following the broader idea of embedding-based cache retrieval \citep{bang2023gptcache}. GenCache generates reusable program caches for structurally similar requests \citep{chakraborty2025generative}. CacheSpec also uses reusable program caches, but differs by parameterizing cached programs and using a small model to extract semantic variables on the cache-hit path.

For methods that involve target-LLM generation, we additionally evaluate +SpecDec variants based on speculative drafting \citep{leviathan2023fast,chen2023accelerating}. In CacheSpec + SpecDec, the same small model serves two roles: it extracts semantic variables for cache-hit requests and acts as the speculative drafter during target-LLM generation. For Formula and CodeTAT-QA, cache-miss requests use code generation and execution, and successful generated programs can become candidates for future cache construction.

\subsection{GenCache Reproduction and Stability}
\label{app:gencache_reproduction}

We use GenCache as the closest program-cache baseline and implement it as a faithful reproduction of the original pipeline \citep{chakraborty2025generative}. The reproduced baseline preserves the original regex/matcher-based variable extraction, request grouping, program-cache generation, cache validation, and fallback execution logic. For each task, we only adapt the input/output interface so that GenCache receives the same task prompts and is evaluated with the same response format as the other methods. ExactCache, GPTCache, GenCache, and CacheSpec are all served through the same API wrapper and evaluated with the same request-level logging and correctness scripts.

This task adaptation is necessary because the evaluated tasks differ in output format: Shopping-Full and Shopping-Struct use shopping-style actions or answers, WebShop uses agent observations and actions, and Formula and CodeTAT-QA use program-generation or answer-checking interfaces. These adapters do not change GenCache's core cache construction or variable extraction mechanism. Therefore, the GenCache results should be interpreted as the performance of its original reuse mechanism under these task interfaces, rather than as a weakened baseline.

\begin{table}[t]
\centering
\caption{
Stability of the reproduced GenCache baseline on Shopping-Full.
}
\label{tab:gencache_stability}
\scriptsize
\setlength{\tabcolsep}{3pt}
\renewcommand{\arraystretch}{1.05}
\resizebox{\columnwidth}{!}{
\begin{tabular}{lccccc}
\hline
\textbf{Setting} & \textbf{Runs} & \textbf{Acc.} & \textbf{Hit Acc.} & \textbf{Cache Gen.} & \textbf{Avg. Lat.} \\
\hline
B1, non-SpecDec & 4 & 80.58 & 80.09 & 27.41 & 0.069--0.197 \\
B3, SpecDec & 4 & 58.97 & 57.77 & 26.71 & 0.040--0.087 \\
\hline
\end{tabular}
}
\end{table}

Table~\ref{tab:gencache_stability} summarizes repeated runs of the original GenCache branch on Shopping-Full. In the non-SpecDec setting (B1), cache-hit accuracy remains within 77.68--82.79\%, while overall accuracy ranges from 78.51\% to 83.13\%. In the SpecDec setting (B3), cache-hit accuracy ranges from 42.59\% to 71.25\%, and overall accuracy ranges from 45.18\% to 71.94\%. Thus, although SpecDec reduces latency for the reproduced GenCache branch, it also introduces substantial variance in cache-hit correctness under this setting.

The same reproduced GenCache implementation also struggles to construct stable reusable caches on structurally perturbed or program-execution-oriented tasks. On Shopping-Struct, GenCache obtains only 0.10\% Hit without SpecDec and 1.46\% Hit with SpecDec. On Formula, GenCache reaches high answer accuracy through the fallback code-generation path, but its Hit is 0.00. These results do not indicate that GenCache is an invalid baseline; rather, they show that the original regex/matcher-based reuse mechanism is brittle when prompts are structurally perturbed or when reusable programs require robust semantic variable extraction.

\paragraph{Metrics.}
For Shopping-Full, Shopping-Struct, Formula, and CodeTAT-QA, we use answer accuracy as the quality metric. For WebShop, we report average reward, following the environment's task-level evaluation protocol \citep{yao2022webshop}. Lat. denotes average per-request end-to-end latency in seconds, measured from request submission to response return.

For cache-based methods, Hit denotes the fraction of requests served by cache. Hit Acc. denotes the fraction of correct outputs among cache-hit requests. This metric is important because high cache hit rate alone does not guarantee reliable reuse: a method may frequently retrieve semantically similar but nonequivalent examples, leading to high Hit but low Hit Acc. and poor overall task quality.

For parallel experiments, Throughput denotes the number of completed requests per second. Wall-clock time per request denotes total wall-clock time divided by the number of completed requests and reflects throughput-side system efficiency under concurrent serving. It differs from Lat., which measures the average end-to-end latency of individual requests. For cache warm-up curves, we use the number of processed requests as the x-axis, which makes cache construction and cache-hit dynamics comparable across concurrency settings with different total running times.

\subsection{Full SpecDec and Non-SpecDec Results}
\label{app:full_specdec_results}

Table~\ref{tab:full_specdec_shopping_webshop} reports the complete non-SpecDec and SpecDec results on Shopping-Full, Shopping-Struct, and WebShop. Table~\ref{tab:full_specdec_financial} reports the complete +SpecDec variants on the financial reasoning tasks. The main text shows representative rows to keep the tables compact.

\begin{table*}[t]
\centering
\caption{
Complete non-SpecDec and SpecDec results on Shopping-Full, Shopping-Struct, and WebShop.
}
\label{tab:full_specdec_shopping_webshop}
\footnotesize
\setlength{\tabcolsep}{3pt}
\renewcommand{\arraystretch}{1.08}
\resizebox{\textwidth}{!}{
\begin{tabular}{lccccccccccc}
\hline
\textbf{Method}
& \multicolumn{4}{c}{\textbf{Shopping-Full}}
& \multicolumn{4}{c}{\textbf{Shopping-Struct}}
& \multicolumn{3}{c}{\textbf{WebShop}} \\
\cline{2-5}\cline{6-9}\cline{10-12}
& Acc. & Lat. & Hit & Hit Acc.
& Acc. & Lat. & Hit & Hit Acc.
& Reward & Lat. & Hit \\
\hline
Direct LLM
& 92.71 & 0.915 & -- & --
& 97.93 & 0.913 & -- & --
& 0.5728 & 1.4070 & -- \\
ExactCache
& 92.72 & 0.972 & 0.00 & --
& 97.93 & 0.971 & 0.00 & --
& 0.5832 & 1.4515 & 0.00 \\
GPTCache
& 8.11 & 0.108 & 91.80 & 0.02
& 17.18 & 0.150 & 89.17 & 7.13
& 0.0531 & 0.1942 & 89.33 \\
GenCache
& 78.51 & 0.197 & 94.89 & 77.68
& 96.91 & 3.252 & 0.10 & 0.00
& 0.5600 & 0.7264 & 70.52 \\
CacheSpec
& 91.91 & 0.256 & 96.80 & 91.89
& 96.65 & 0.249 & 98.03 & 96.62
& 0.5763 & 0.6598 & 74.22 \\
Direct LLM + SpecDec
& 92.73 & 0.419 & -- & --
& 97.94 & 0.475 & -- & --
& 0.5762 & 1.1364 & -- \\
ExactCache + SpecDec
& 92.70 & 0.584 & 0.00 & --
& 97.94 & 0.527 & 0.00 & --
& 0.5762 & 1.1629 & 0.00 \\
GPTCache + SpecDec
& 8.11 & 0.073 & 91.80 & 0.02
& 17.16 & 0.096 & 89.17 & 7.11
& 0.0531 & 0.1667 & 89.33 \\
GenCache + SpecDec
& 59.09 & 0.040 & 98.95 & 58.67
& 95.99 & 1.852 & 1.46 & 23.29
& 0.5680 & 0.6298 & 66.40 \\
CacheSpec + SpecDec
& 92.15 & 0.214 & 96.35 & 92.13
& 96.66 & 0.228 & 98.03 & 96.64
& 0.5760 & 0.5797 & 74.22 \\
\hline
\end{tabular}
}
\end{table*}

\begin{table*}[t]
\centering
\caption{
Complete non-SpecDec and SpecDec results on Formula and CodeTAT-QA.
}
\label{tab:full_specdec_financial}
\footnotesize
\setlength{\tabcolsep}{3pt}
\renewcommand{\arraystretch}{1.08}
\begin{tabular}{lcccccccc}
\hline
\textbf{Method}
& \multicolumn{4}{c}{\textbf{Formula}}
& \multicolumn{4}{c}{\textbf{CodeTAT-QA}} \\
\cline{2-5}\cline{6-9}
& Acc. & Lat. & Hit & Hit Acc.
& Acc. & Lat. & Hit & Hit Acc. \\
\hline
Direct LLM
& 75.92 & 0.237 & -- & --
& 49.17 & 0.258 & -- & -- \\
PoT-style
& 94.69 & 2.015 & -- & --
& 71.53 & 2.447 & -- & -- \\
ExactCache
& 94.64 & 2.056 & 0.00 & --
& 71.98 & 2.550 & 0.08 & 100.00 \\
GPTCache
& 94.56 & 0.858 & 63.45 & 93.69
& 42.54 & 0.485 & 83.03 & 35.73 \\
GenCache
& 94.52 & 3.678 & 0.00 & --
& 71.69 & 3.232 & 0.00 & -- \\
CacheSpec
& 94.19 & 0.847 & 84.14 & 93.54
& 71.25 & 1.849 & 56.05 & 68.00 \\
Direct LLM + SpecDec
& 75.80 & 0.314 & -- & --
& 49.24 & 0.472 & -- & -- \\
PoT-style + SpecDec
& 94.56 & 1.564 & -- & --
& 71.69 & 1.880 & -- & -- \\
ExactCache + SpecDec
& 94.64 & 1.650 & 0.00 & --
& 72.03 & 1.720 & 0.08 & 100.00 \\
GPTCache + SpecDec
& 94.56 & 0.700 & 63.45 & 93.69
& 42.36 & 0.331 & 83.03 & 35.57 \\
GenCache + SpecDec
& 94.56 & 2.873 & 0.00 & --
& 71.67 & 3.042 & 0.00 & -- \\
CacheSpec + SpecDec
& 94.19 & 0.648 & 87.16 & 93.76
& 70.41 & 1.648 & 57.48 & 71.79 \\
\hline
\end{tabular}
\end{table*}

\subsection{Detailed Main Result Analysis}
\label{app:detailed_main_results}

On Shopping-Full, CacheSpec substantially reduces latency while maintaining task quality close to Direct LLM. With SpecDec, the framework reaches 92.15 Acc. and 0.214s Lat., compared with 92.71 Acc. and 0.915s Lat. for Direct LLM. Thus, the result should not be interpreted as improving the target LLM's intrinsic accuracy. Instead, the key evidence is that most requests are served by cache (96.35 Hit) and that these cache-hit outputs remain reliable (92.13 Hit Acc.), indicating that reusable programs can replace repeated target-LLM generation with only a small quality difference on Shopping-Full.

Shopping-Struct stresses robustness under structural perturbations. GenCache obtains high task quality on this dataset, but its Hit is only 0.10, indicating that it mainly falls back to target-LLM generation rather than reusing cache entries. In contrast, CacheSpec with SpecDec achieves 98.03 Hit and 96.64 Hit Acc., while reducing Lat. from Direct LLM's 0.913s to 0.228s. This comparison directly supports the role of semantic variable extraction: it allows the system to reuse cached programs even when requests preserve semantic intent but change surface structure.

GPTCache serves as an instructive counterexample. It reaches high Hit on Shopping-Full and Shopping-Struct, but its task quality collapses. This shows that full-prompt semantic similarity alone is insufficient for safe reuse: similar prompts may still require different variable bindings or different actions. CacheSpec avoids this failure mode by extracting task variables before executing a cached program, which leads to high Hit and high Hit Acc. simultaneously.

On WebShop, CacheSpec maintains the same level of task reward as Direct LLM while reducing latency. With SpecDec, the framework reaches 0.5760 average reward with 0.5797s latency, compared with 0.5728 reward and 1.4070s latency for Direct LLM. GPTCache again shows that high cache hit rate is not enough: despite 89.33 Hit, its reward drops to 0.0531. The identical Hit of CacheSpec and CacheSpec + SpecDec is expected because SpecDec changes the target-LLM generation path but not the cache matching policy. ExactCache obtains a slightly higher reward in this 50-episode sample, but its Hit is 0.00; therefore, the result reflects fallback to the Direct LLM path rather than effective exact-cache reuse.

On Formula, PoT-style substantially improves accuracy over Direct LLM, increasing Acc. from 75.92 to 94.69, but it also increases latency from 0.237s to 2.015s. CacheSpec preserves PoT-style-level accuracy while reducing latency through program reuse. CacheSpec + SpecDec achieves 94.19 Acc. with 0.648s latency, giving a large reduction compared with PoT-style while maintaining 87.16 Hit and 93.76 Hit Acc. This indicates that cached programs can serve as reusable reasoning artifacts for formula computation tasks.

The comparison with GenCache is central. On Formula, GenCache reaches 94.52 Acc., but its Hit is 0.00. Thus, its accuracy mainly comes from the fallback code-generation path rather than from cache reuse. CacheSpec, by contrast, achieves high Hit and high Hit Acc., showing that small-model semantic variable extraction makes the program cache executable for future requests. A similar pattern appears on CodeTAT-QA: GenCache again has 0.00 Hit, while CacheSpec + SpecDec reaches 57.48 Hit and 71.79 Hit Acc.

CodeTAT-QA is more challenging than Formula because it involves table-and-text reasoning and weaker group-level regularities. CacheSpec + SpecDec obtains slightly lower accuracy than PoT-style + SpecDec, but reduces latency from 1.880s to 1.648s while serving more than half of the requests through cache. The slight accuracy drop on CodeTAT-QA suggests that program caching requires sufficiently stable group-level computation patterns to fully preserve PoT-style performance.

SpecDec is useful mainly when target-LLM generation is long enough to amortize draft-and-verify overhead. In the direct-output setting, Direct LLM + SpecDec can be slower than Direct LLM: on Formula, latency increases from 0.237s to 0.314s, and on CodeTAT-QA, it increases from 0.258s to 0.472s. In contrast, PoT-style + SpecDec reduces latency compared with PoT-style, and CacheSpec + SpecDec further reduces latency compared with CacheSpec. This supports our design choice of reusing the small model as a speculative drafter for code generation and cache construction, rather than treating SpecDec as a universal speedup for every output length.

\subsection{Parallel Performance Details}
\label{app:parallel_performance}

As shown in panels (a) and (b) of Figure~\ref{fig:parallel_indicators}, CacheSpec achieves higher throughput and lower wall-clock time per request across all concurrency levels. At concurrency 16, CacheSpec reaches its peak throughput of 18.56 req/s, compared with 6.51 req/s for PoT-style. At the same concurrency level, it reduces wall-clock time per request to 0.0539s, while PoT-style requires 0.1535s. When concurrency increases to 32, its throughput slightly drops to 17.38 req/s and wall-clock time per request increases to 0.0575s, suggesting that excessive concurrency begins to introduce queueing or resource contention.

The parallel advantage comes from replacing repeated target-LLM program generation with the cache-hit path. After a reusable program cache is available, CacheSpec only needs to extract variables and execute the cached program for later requests in the same group. Panel (c) illustrates this online cache warm-up process with a representative run at concurrency 8. The number of cache entries grows quickly during approximately the first 1K processed requests and reaches 25 entries around 1.1K requests, after which it stabilizes. The cumulative hit rate increases accordingly, and the final hit rate of the full run reaches approximately 98.23\%.

These results highlight an important trade-off in parallel serving. Moderate concurrency improves throughput and provides enough requests for cache generation, while excessive concurrency may send more requests to the miss path before useful caches are available, increasing target-LLM pressure. Thus, the parallel efficiency of CacheSpec depends not only on the execution cost of the cache-hit path, but also on the match between cache generation speed and request arrival rate.

\subsection{Length Robustness Details}
\label{app:length_robustness}

As the average context length increases from Avg-1K to Avg-8K, the latency of PoT-style increases from 2.577s to 5.737s, while the latency of CacheSpec increases from 0.574s to 2.443s. CacheSpec consistently outperforms PoT-style across all length settings, with speedups of 4.49$\times$, 4.11$\times$, 3.77$\times$, and 2.35$\times$, respectively. The reduced speedup at Avg-8K suggests that very long contexts increase the overall cost of variable extraction, cache-hit execution, and miss/fallback paths, but the remaining 2.35$\times$ speedup still shows that program caching is effective for long-context formula computation.

CacheSpec also preserves stable quality under long contexts. From Avg-1K to Avg-4K, its accuracy, cache hit rate, and cache-hit accuracy remain around or above 94\%. At Avg-8K, accuracy decreases to 93.01\%, cache hit rate decreases to 90.85\%, and cache-hit accuracy remains 93.34\%. This indicates that extremely long contexts make cache reuse more difficult, but they do not cause a collapse in either answer quality or cache-hit correctness. The combination of high end-to-end accuracy and high cache hit rate further indicates that cache-hit requests are also reliable, so cached programs are both frequently used and correct when applied. These results suggest that when tasks have stable formula-level computation structures, reusable program caches can still provide substantial latency reduction under long-context inputs.

\section{Additional Revision Experiments}
\label{app:revision_experiments}

This section reports additional experiments conducted during revision. They examine model choice, repeated-run stability, cold-start behavior, semantic-routing sensitivity, semantic variable extraction, and cache-error propagation. Unless stated otherwise, the additional experiments use strictly sequential request processing, temperature 0, and disabled thinking mode. Negative Hit denotes the percentage of cache-hit requests judged incorrect.

\subsection{Small-Model Choice and Cross-Family Transfer}
\label{app:model_choice}

We first vary the small model on Shopping-Struct. Within the Qwen family, Qwen3-0.6B, Qwen3-1.7B, and Qwen3-4B are paired with a fixed Qwen3-32B target LLM and judge. We also test transfer to another model family by pairing Llama-3.2-1B and Llama-3.2-3B with Llama-3.3-70B. In every CacheSpec configuration, the small model serves as both the semantic variable extractor and the SpecDec drafter.

\begin{table*}[t]
\centering
\caption{Model-choice results on Shopping-Struct. Acc., Hit, Hit Acc., and Negative Hit are percentages; Lat. is average request latency in seconds. Mean $\pm$ standard deviation is computed over three runs where available.}
\label{tab:revision_model_choice}
\footnotesize
\setlength{\tabcolsep}{3.5pt}
\renewcommand{\arraystretch}{1.08}
\resizebox{\textwidth}{!}{
\begin{tabular}{lcccccc}
\hline
\textbf{Model setting} & \textbf{Runs} & \textbf{Acc.} & \textbf{Hit} & \textbf{Hit Acc.} & \textbf{Negative Hit} & \textbf{Lat. (s)} \\
\hline
CacheSpec: Qwen3-0.6B + Qwen3-32B & 3 & $97.27\!\pm\!0.015$ & 97.46 & 97.22 & 2.78 & 0.1216 \\
CacheSpec: Qwen3-1.7B + Qwen3-32B & 3 & $99.47\!\pm\!0.025$ & 98.07 & 99.47 & 0.53 & 0.1553 \\
CacheSpec: Qwen3-4B + Qwen3-32B & 3 & $99.83\!\pm\!0.006$ & 98.13 & 99.83 & 0.17 & 0.2254 \\
Direct Qwen3-32B & 1 & 99.94 & -- & -- & -- & 2.5581 \\
CacheSpec: Llama-3.2-1B + Llama-3.3-70B & 3 & $98.70\!\pm\!0.026$ & 99.17 & 98.73 & 1.27 & 0.0800 \\
CacheSpec: Llama-3.2-3B + Llama-3.3-70B & 3 & $98.33\!\pm\!0.026$ & 99.50 & 98.36 & 1.64 & 0.1600 \\
Direct Llama-3.3-70B & 3 & $99.06\!\pm\!0.006$ & -- & -- & -- & 1.2200 \\
\hline
\end{tabular}
}
\end{table*}

The Qwen results reveal a reliability--cost trade-off. Increasing the small-model size reduces Negative Hit from 2.78\% to 0.17\%, but increases latency from 0.1216s to 0.2254s. All three CacheSpec variants remain substantially faster than direct Qwen3-32B inference. The Llama results show that the mechanism transfers beyond the Qwen family and retains a clear latency advantage. However, the Llama 1B/3B comparison is not monotonic in accuracy, so the ``sweet spot'' should not be interpreted as a universal function of parameter count. It depends jointly on extraction reliability, task structure, draft-model overhead, and target-model latency. The small standard deviations under deterministic sequential serving also indicate that the additional results are stable across repeated runs.

\subsection{Cold-Start and Data-Scale Behavior}
\label{app:cold_start_scale}

To explain the high cache-hit rates, we randomly shuffle Formula and evaluate increasingly large prefixes, with every scale starting from an empty cache. Table~\ref{tab:formula_scale_growth} shows that Hit rises as the request stream grows and the cold-start portion becomes smaller. This pattern reflects reuse of intermediate computation structures rather than exact prompt repetition or final-answer reuse.

\begin{table}[t]
\centering
\caption{Cold-start scale results on Formula. Acc., Hit, Hit Acc., and Negative Hit are percentages; Lat. is average request latency in seconds.}
\label{tab:formula_scale_growth}
\scriptsize
\setlength{\tabcolsep}{3pt}
\renewcommand{\arraystretch}{1.05}
\resizebox{\columnwidth}{!}{
\begin{tabular}{rccccc}
\hline
\textbf{$N$} & \textbf{Acc.} & \textbf{Hit} & \textbf{Hit Acc.} & \textbf{Negative Hit} & \textbf{Lat. (s)} \\
\hline
100  & 94.00 & 25.00 & 96.00 & 4.00 & 3.4445 \\
250  & 92.80 & 63.60 & 92.45 & 7.55 & 2.3871 \\
500  & 93.20 & 77.80 & 94.86 & 5.14 & 1.2749 \\
1000 & 92.90 & 87.30 & 94.39 & 5.61 & 0.6674 \\
1500 & 89.80 & 90.80 & 89.87 & 10.13 & 0.5604 \\
2000 & 92.30 & 90.85 & 93.29 & 6.71 & 0.5498 \\
2446 & 92.85 & 92.93 & 94.06 & 5.94 & 0.4285 \\
\hline
\end{tabular}
}
\end{table}

As $N$ increases from 100 to 2,446, Hit rises from 25.00\% to 92.93\% and average latency falls from 3.4445s to 0.4285s. Hit Acc. is not monotonic, however, and drops to 89.87\% at $N=1{,}500$. This result reinforces that high Hit alone is insufficient: the cache-hit outputs must also be validated for correctness.

\subsection{Routing Sensitivity and Cache Settings}
\label{app:routing_sensitivity}

Table~\ref{tab:revision_hyperparameters} reports the key settings used for the revision experiments. Both Shopping-Struct and Formula use a default semantic threshold of 0.85. Cache generation is attempted once a semantic group accumulates three requests, and a candidate is accepted when at least two of three validation requests pass.

\begin{table}[t]
\centering
\caption{Key hyperparameters used in the revision experiments.}
\label{tab:revision_hyperparameters}
\scriptsize
\setlength{\tabcolsep}{3pt}
\renewcommand{\arraystretch}{1.08}
\resizebox{\columnwidth}{!}{
\begin{tabular}{lcccc}
\hline
\textbf{Task} & \textbf{Grouping} & \textbf{$\tau$} & \textbf{Trigger} & \textbf{Validation} \\
\hline
Shopping-Struct & Semantic + schema & 0.85 & 3 requests & 2 of 3 pass \\
Formula & Semantic & 0.85 & 3 requests & 2 of 3 pass \\
\hline
\end{tabular}
}
\end{table}

We sweep the semantic routing threshold on Shopping-Struct while keeping the remaining settings fixed. As shown in Table~\ref{tab:semantic_threshold_sweep}, thresholds from 0.50 to 0.90 preserve high accuracy, but higher thresholds route more requests to fallback and reduce reuse. At 0.92, Hit falls sharply and Hit Acc. also degrades, showing that excessively conservative routing can alter grouping and cache-construction behavior rather than guaranteeing better accuracy.

\begin{table}[t]
\centering
\caption{Semantic-threshold sensitivity on Shopping-Struct. Acc., Hit, Hit Acc., and Negative Hit are percentages; Fallback is the number of target-LLM fallback calls.}
\label{tab:semantic_threshold_sweep}
\scriptsize
\setlength{\tabcolsep}{2.5pt}
\renewcommand{\arraystretch}{1.04}
\resizebox{\columnwidth}{!}{
\begin{tabular}{cccccc}
\hline
\textbf{$\tau$} & \textbf{Acc.} & \textbf{Hit} & \textbf{Hit Acc.} & \textbf{Negative Hit} & \textbf{Fallback} \\
\hline
0.50 & 99.45 & 99.70 & 99.45 & 0.55 & 30 \\
0.60 & 99.44 & 99.68 & 99.44 & 0.56 & 32 \\
0.70 & 99.43 & 99.62 & 99.43 & 0.57 & 38 \\
0.80 & 99.46 & 99.31 & 99.46 & 0.54 & 69 \\
0.82 & 99.48 & 99.13 & 99.48 & 0.52 & 87 \\
0.85 & 99.47 & 98.07 & 99.47 & 0.53 & 193 \\
0.88 & 99.59 & 94.75 & 99.58 & 0.42 & 525 \\
0.90 & 99.58 & 89.77 & 99.54 & 0.46 & 1023 \\
0.92 & 98.18 & 64.31 & 97.23 & 2.77 & 3569 \\
\hline
\end{tabular}
}
\end{table}

These parameters expose a practical trade-off. A larger cache trigger provides more same-group observations before generation and may improve template stability, but increases cold-start cost. A stricter validation rule can reduce erroneous-cache risk, but also lowers cache-generation success and future Hit. Likewise, increasing $\tau$ makes routing more conservative but can substantially reduce the latency benefit.

\subsection{Semantic Variable Extraction Analysis}
\label{app:variable_extraction_analysis}

For Formula, extraction references are constructed from each generated cache's variable specification and program code. A separate strong model produces pseudo-gold slot values, and only references whose program execution matches the dataset ground truth are retained. We measure per-slot accuracy, the percentage of requests with all slots correct, and execution-equivalent accuracy, which checks whether the extracted values produce the same program output as the reference.

\begin{table}[t]
\centering
\caption{Formula extraction accuracy across input lengths. All metrics are percentages. Mean prompt lengths for Avg-256 and Avg-512 are 256.17 and 512.27 tokens.}
\label{tab:extraction_by_length}
\scriptsize
\setlength{\tabcolsep}{4pt}
\renewcommand{\arraystretch}{1.05}
\resizebox{\columnwidth}{!}{
\begin{tabular}{lccc}
\hline
\textbf{Input length} & \textbf{Slot Acc.} & \textbf{All-Slots Acc.} & \textbf{Execution-Equivalent Acc.} \\
\hline
Avg-256 & 97.68 & 97.49 & 97.80 \\
Avg-512 & 97.68 & 97.49 & 97.90 \\
Avg-1K  & 97.63 & 97.39 & 98.10 \\
Avg-2K  & 97.73 & 97.49 & 98.10 \\
Avg-4K  & 97.73 & 97.49 & 98.10 \\
Avg-8K  & 97.73 & 97.49 & 98.10 \\
\hline
\end{tabular}
}
\end{table}

The Formula results remain stable from Avg-256 to Avg-8K. Table~\ref{tab:extraction_by_count_type} further analyzes the number and type of input variables. The current benchmark mainly contains two or three program input slots, so the variable-count result should not be generalized to programs with many variables. Structured numeric variables are generally stable, with remaining risks concentrated in amount scaling, units, and percentage-versus-decimal conventions.

\begin{table}[t]
\centering
\caption{Formula extraction accuracy by variable count and type. All values are percentages. Counts exclude outputs and program-internal derived variables.}
\label{tab:extraction_by_count_type}
\scriptsize
\setlength{\tabcolsep}{5pt}
\renewcommand{\arraystretch}{1.05}
\begin{tabular}{lcc}
\hline
\textbf{Category} & \textbf{Slot Acc.} & \textbf{All-Slots Acc.} \\
\hline
2 input variables & 96.97 & 96.84 \\
3 input variables & 99.72 & 99.16 \\
\hline
Amount & 96.70 & -- \\
Count or period & 97.92 & -- \\
Other numeric & 98.80 & -- \\
Price or share & 98.09 & -- \\
Rate or ratio & 100.00 & -- \\
\hline
\end{tabular}
\end{table}

Shopping-Struct is more difficult under strict field-level evaluation because the extractor must recover free-form product keywords and exact constraints. The all-slots accuracy is 29.30\%, keyword token F1 is 83.91\% (82.77\% precision and 87.23\% recall), and exact maximum-price accuracy is 40.80\%. These extraction metrics are stricter than final action-level task accuracy and expose errors that can be hidden by semantically acceptable final actions.

\begin{table}[t]
\centering
\caption{Main semantic-variable extraction failure modes. Counts are numbers of observed errors in the revision analysis.}
\label{tab:extraction_failures}
\scriptsize
\setlength{\tabcolsep}{3pt}
\renewcommand{\arraystretch}{1.05}
\resizebox{\columnwidth}{!}{
\begin{tabular}{llrp{0.47\columnwidth}}
\hline
\textbf{Task} & \textbf{Failure type} & \textbf{Count} & \textbf{Explanation} \\
\hline
Formula & Slot-value error & 101 & Amount scaling, unit abbreviation, or numeric binding. \\
Formula & Execution mismatch & 56 & Extracted values produce an output inconsistent with ground truth. \\
Formula & Percent-scale difference & 15 & Percentage and decimal-scale mismatch. \\
Shopping-Struct & Price error & 587 & Exact price constraints are incorrectly recovered. \\
Shopping-Struct & Keyword hallucination & 332 & Extracted attributes are absent from the gold keywords. \\
Shopping-Struct & Keyword omission & 245 & Important product attributes or constraints are missing. \\
\hline
\end{tabular}
}
\end{table}

\subsection{Cache-Error Propagation}
\label{app:cache_error_analysis}

The full Formula scale experiment produces 175 incorrect outputs: 135 on cache-hit paths and 40 on cache-miss or fallback paths. Cache-hit errors include cached-program mismatch and variable-binding errors, whereas miss-path errors are produced without executing a cached program. Thus, although cache reuse provides the main latency benefit, an incorrect cache can systematically propagate errors across later requests.

\begin{table*}[t]
\centering
\caption{Templates with the highest error counts in the full Formula scale experiment. Errors and Cache Errors are counts; In-Group Hit is a percentage.}
\label{tab:cache_error_templates}
\footnotesize
\setlength{\tabcolsep}{4pt}
\renewcommand{\arraystretch}{1.06}
\resizebox{\textwidth}{!}{
\begin{tabular}{lrrcl}
\hline
\textbf{Template} & \textbf{Errors} & \textbf{Cache Errors} & \textbf{In-Group Hit (\%)} & \textbf{Main cause} \\
\hline
Capital Asset Pricing Model (CAPM) & 91 & 62 & 66.7 & Answer-scale inconsistency, e.g., 0.05 versus 5.3. \\
Modigliani--Miller Theorem & 23 & 23 & 97.6 & Tax-rate percentage/decimal scale and amount binding. \\
Days Sales Outstanding (DSO) & 14 & 14 & 97.1 & Numeric rounding-format differences. \\
Inventory Turnover & 14 & 14 & 96.9 & Million-scale versus raw-amount unit mismatch. \\
Price-to-Earnings Ratio (P/E) & 14 & 14 & 97.3 & Rounding-precision differences. \\
\hline
\end{tabular}
}
\end{table*}

CAPM is the dominant failure source because examples within the same apparent group use inconsistent answer scales. Similar problems arise from percentage/decimal conventions, unit scaling, and rounding requirements. These results identify an applicability boundary: workloads must have not only repeated structure but also consistent input and output conventions. Stronger cache validation, finer-grained grouping, explicit unit normalization, and answer-scale consistency checks are promising safeguards against systematic error propagation.

\end{document}